\pdfoutput=1

\documentclass[11pt]{article}

\usepackage[preprint]{acl}

\usepackage{newtx}
\usepackage{latexsym}

\usepackage{amsmath}
\usepackage{amsfonts}
\usepackage{algorithm}
\usepackage{algpseudocode}
\usepackage{pseudocode}
\usepackage{multirow}
\usepackage{colortbl}
\usepackage{color}
\usepackage{epigraph}
\usepackage{caption}
\usepackage{subcaption}
\usepackage{hyperref}
\usepackage{tabularx}
\usepackage{float}
\usepackage{longtable}
\usepackage[pdftex]{graphicx}
\usepackage{pdfpages}
\usepackage{pdflscape}
\usepackage[acronym,toc]{glossaries}
\usepackage{setspace}

\usepackage{subcaption}
\usepackage{acronym}
\usepackage{booktabs}
\usepackage{listings}
\usepackage{svg}

\usepackage{fancyvrb} 

\usepackage{CJKutf8}

\usepackage{fancyvrb}
\usepackage{fvextra} 

\definecolor{darkgreen}{rgb}{0.0, 0.5, 0.0}
\definecolor{darkred}{rgb}{0.8, 0.0, 0.0}

\newcommand{\up}[1]{{#1 \textcolor{darkgreen}{$\uparrow$}}}
\newcommand{\down}[1]{{#1 \textcolor{darkred}{$\downarrow$}}}

\usepackage{amsmath}
\DeclareMathOperator*{\argmax}{arg\,max}

\usepackage[T1]{fontenc}

\usepackage[utf8]{inputenc}

\usepackage{microtype}

\usepackage{inconsolata}

\usepackage{graphicx}

\newacronym{LLM}{LLM}{Large Language Model}
\newacronym{DAG}{DAG}{Directed Acyclic Graph}
\newacronym{NLP}{NLP}{Natural Language Processing}
\newacronym{MoE}{MoE}{Mixture of Experts}
\newacronym{GPT}{GPT}{Generative Pre-Trained Transformer}
\newacronym{MMLU}{MMLU}{Massive Multitask Language Understanding}
\newacronym{CMMLU}{CMMLU}{Chinese Massive Multitask Language Understanding}
\newacronym{CoT}{CoT}{Chain-of-Thought}
\newacronym{RL}{RL}{Reinforcement Learning}

\title{Input Conditioned Graph Generation for Language Agents}

\author{
    Lukas Vierling, Jie Fu\footnotemark[1], Kai Chen\footnotemark[1] \\
    Hong Kong University of Science and Technology \\
    \texttt{lvierling@connect.ust.hk, jiefu@ust.hk, kaichen@cse.ust.hk}
}

\begin{document}
\maketitle
\begin{abstract}
Recent progress in Large Language Models (LLMs) and language agents has demonstrated significant promise for various future applications across multiple disciplines. While traditional approaches to language agents often rely on fixed, handcrafted designs, our research aims to develop both learnable and dynamic agents.
Our method uses an existing framework that abstracts language agents as graphs. Within this graph framework, we aim to learn a model that can generate edges for every given input to the language agent. This allows us to generate edges that represent the flow of communication within the graph based on the given input, thereby adjusting the internal communication of a language agent. We learn to generate these edges using a pretrained LLM that is fine-tuned with reinforcement learning. This LLM can be fine-tuned on several datasets simultaneously, and we hypothesize that the model learns to adapt to these different domains during training, achieving good overall performance when encountering data from different domains during deployment. We demonstrate that our approach surpasses the previous static approach by nearly 6\% accuracy on a combined dataset of MMLU and CMMLU, and by more than 10\% when trained with a sparsity-inducing loss. It also performs superior in additional experiments conducted with the MMLU and Mini Crossword Puzzles datasets. The code is available at \url{https://github.com/lukasVierling/DynamicGPTSwarm}. 
\end{abstract}

\renewcommand{\thefootnote}{\fnsymbol{footnote}}
\footnotetext[1]{Corresponding authors}
\renewcommand{\thefootnote}{\arabic{footnote}}

\section{Introduction}
\label{Introduction}
Recent advancements in \glspl{LLM} have significantly expanded their potential applications. Pretrained \glspl{LLM} can effectively handle a wide range of \gls{NLP} tasks with little to no additional training, a capability known as zero-shot or few-shot learning \cite{brown2020language,touvron2023llama,team2024gemma}. This enables their use in various critical applications, either to solve complex problems or to support human workflows \cite{chen2023meditron,colombo2024saullm,roziere2023code,xu2023paradigm}.

A notable application of \glspl{LLM} is the development of language agents. These agents use \glspl{LLM} as their core component and perform various tasks through interactions among multiple \glspl{LLM}, enhanced by additional operations such as memory retrieval, code execution, and environmental interactions like search \cite{birr2024autogpt+,hong2023metagpt,Chase_LangChain_2022}. In contrast to usual \glspl{LLM}, language agents interact with various environments by leveraging \glspl{LLM} through actions and observations. Unlike \glspl{LLM}, language agents can perform internal actions such as reasoning, which may involve multiple \gls{LLM} queries before interacting with the environment. They also employ various tools to engage with data sources. Language agents can function in single-agent or multi-agent frameworks, using one or more \glspl{LLM} \cite{wang2024survey}. These agents can be trained using \gls{RL} techniques \cite{zhuge2024language}, although much of the existing literature focuses on handcrafted designs that utilize pretrained \glspl{LLM} without further training \cite{hong2023metagpt, Chase_LangChain_2022, chen2023agentverse}.

\subsection{Towards Learned and Dynamic Language Agents}

Over the past few decades of deep learning research, a recurring pattern has been the superiority of learned features over handcrafted ones \cite{mikolov2013distributed,hinton2012deep,silver2017mastering}. Notable examples include AlexNet by \citet{krizhevsky2012imagenet}, which significantly outperformed the state-of-the-art on ImageNet by employing Convolutional Neural Networks, and Neural Architecture Search algorithms, which improved performance across various benchmarks through architectures discovered via automated searches rather than handcrafted designs \cite{zoph2016neural}. While it remains a topic of debate whether learned approaches consistently surpass handcrafted designs, ample evidence suggests their potential for superiority \cite{krizhevsky2012imagenet,mikolov2013distributed,hinton2012deep,silver2017mastering,zoph2016neural}.

In the realm of language agents, many existing approaches incorporate handcrafted designs \cite{hong2023metagpt,Chase_LangChain_2022,chen2023agentverse,Liu_LlamaIndex_2022,zhuge2023mindstorms}, often tailored explicitly for specific tasks. For instance, the MetaGPT framework assigns predefined roles to agents, mimicking human workflows \cite{hong2023metagpt}. This strategy has shown promise, particularly in coding benchmarks, but it also introduces significant inductive biases by imposing human-like workflows on language agents, which may limit their potential by constraining their design.

Another important ability is adaptability to input variations, allowing systems to manage different types of data through distinct processing steps. Research on \gls{CoT} prompting has highlighted its advantages for mathematical or reasoning tasks, where several reasoning steps precede the final output \cite{wei2022chain}. Similarly, the Tree of Thought approach has demonstrated improved performance in tasks such as crossword puzzles by exploring various potential answers \cite{yao2024tree}. Based on this, we hypothesize that applying different strategies or workflows based on the given input can optimize task solutions. A one-size-fits-all solution may serve as a starting point, but over the long term, language agents should have the flexibility to explore various communication flows and apply tailored methods to enhance their performance.

Our work builds upon the framework proposed by \citet{zhuge2024language}, which models language agents as Directed Acyclic Graphs (DAGs). This framework allows for an abstract understanding of language agents by representing them as computational graphs where nodes perform specific operations and edges depict the flow of data. We extend this DAG-based approach by introducing adaptive language agents that can modify their internal and external communications based on initial input. Using reinforcement learning, specifically the REINFORCE algorithm \cite{williams1992simple}, we aim to optimize the communication flows within these agents. Unlike previous methods with fixed edge probabilities, our approach learns input-dependent edge probabilities by utilizing a \gls{LLM}, allowing for dynamic and context-sensitive graph structures.

We aim to assess the performance of our method through three primary experiments using the Crosswords Puzzle dataset \cite{yao2024tree}. These experiments evaluate the capability of language agents in solving 5x5 crossword puzzles, with performance measured by the number of correctly predicted words. The second experiment uses the \gls{MMLU} \citep{hendrycks2021measuring} dataset for question answering to evaluate reasoning capabilities and detect adversarial agents within the graph. The final experiment combines the \gls{MMLU} and \gls{CMMLU} \citep{li2023cmmlu} datasets to test our method's ability to handle inputs from diverse domains, with performance measured by the number of correctly answered questions.

Our contributions are summarized as follows:
\begin{itemize}
    \item We propose a novel method for edge optimization in language agents, enabling input-dependent graph generation.
    \item We provide theoretical justification and demonstrate the superiority of our method through experimental validation.
\end{itemize}

\section{Related Work}
\label{Related Work}
Recent language models like GPT-3 \cite{brown2020language}, LLama \cite{touvron2023llama2}, and Claude\footnote{\url{https://docs.anthropic.com/claude/docs/models-overview}} excel in diverse NLP tasks through unified architectures \cite{achiam2023gpt, jiang2023mistral, team2024gemma, touvron2023llama}. Extensive pretraining allows for zero-shot and few-shot prompting without task-specific fine-tuning \cite{brown2020language}. Few-shot prompting uses example pairs for contextual learning, while zero-shot relies solely on task descriptions. Our research extends this by using language agents that interact with their environment. This approach enhances \gls{LLM} functionality, enabling complex tasks like reasoning and memory retrieval by interacting with external tools and data sources \cite{Liu_LlamaIndex_2022, Chase_LangChain_2022, birr2024autogpt+, reed2022generalist}.
\subsection{Language Agents in Role Play Setting}
In the existing literature, considerable attention has been devoted to exploring the potential of assigning specific roles to language agents to enhance their problem-solving capabilities \cite{zhuge2023mindstorms,li2024camel,hong2023metagpt,qian2023communicative}. NLSOM by \citet{zhuge2023mindstorms} employs a society of mind concept, where multiple \glspl{LLM} and neural network-based experts operate within a structured society, exchanging information to facilitate complex decisions. Agents in NLSOM follow predefined roles akin to political systems, such as democracies and monarchies, which, while structured, lack flexibility and require handcrafted organizational structures. Similarly, the CAMEL framework by \citet{li2024camel} assigns specific roles to agents to guide problem-solving processes, emphasizing role adherence to enhance creativity. However, CAMEL’s predefined role assignments and fixed communication schemes limit its adaptability. MetaGPT by \citet{hong2023metagpt} focuses on role specialization and improved communication infrastructure, relying on user-defined roles and human workflow patterns adapted from software engineering. It uses a subscribe-and-publish mechanism for agent communication, offering some flexibility but still constrained by predefined workflows. In contrast, our research introduces a dynamic approach to inter-agent communication using reinforcement learning techniques, enabling agents to learn and adapt their communication strategies over time. This flexibility allows agents to modify their internal communication based on real-time task requirements and performance, leading to more robust and adaptable problem-solving capabilities, enhancing responsiveness to complex and evolving tasks.
\subsection{Dynamic Language Agents}
Although various methodologies have been developed for dynamically generating language agents tailored to specific task requirements \cite{xagent2023,chen2023agentverse,liu2023dynamic,yao2022react}, most focus on role assignment and task-dependent agent creation. In contrast, our approach utilizes a fixed set of agents and concentrates on optimizing their communication. XAgent by \citet{xagent2023} is an open-source framework with a Dispatcher, Planner, and Actor, relying on \glspl{LLM} for planning and dispatching tasks. Our method, however, integrates these functions within graph edge generation, using reinforcement learning for improved task-handling strategies. AgentVerse by \citet{chen2023agentverse} employs a multi-stage problem-solving process with dynamic expert recruitment and structured decision-making, while our approach autonomously learns decision-making procedures using a utility function for feedback, allowing task transferability. DyLAN by \citet{liu2023dynamic} uses inference-time agent selection based on an Agent Importance Score, restricted to multi-round interactions. Our method, instead, employs a generalized graph framework without limiting inter-agent connections, evaluated through a utility function on a dataset. Overall, our approach focuses on an abstracted graph framework that optimizes internal communication between agents using \gls{RL} techniques, differing from methods that rely on \glspl{LLM}' internal knowledge for decision-making.

\section{Methodology}
\label{Methodology}

\subsection{Language Agents as Graphs}

Language agents, enhanced by pretrained \glspl{LLM}, have shown significant promise in leveraging their extensive knowledge to handle various tasks. These agents often adopt complex behaviors, such as teamwork and role assignments, to improve task performance \cite{hong2023metagpt,chen2023agentverse,qian2023communicative}. The framework proposed by \citet{zhuge2024language} models language agents as Directed Acyclic Graphs (DAGs). In this approach, a language agent is defined as a graph \(G(V, E, F, o)\), where \(V\) is the set of nodes, \(E\) is the set of edges between these nodes representing the flow of data within the language agent, \(F\) is a set of operations with \(f_i\) being the operation executed in node \(v_i\), and \(o\) is the output node. In their research, they put special emphasis on language agent swarms. These swarms are graphs composed of several subgraphs, where every subgraph represents a single language agent. For their edge optimization techniques, they restricted the optimization to the edges between different agents within the composed graph. This restricted set of edges is \(\mathcal{E} \subset E\).

\subsection{Static Edge Probabilities}

A language agent as a graph can be executed based on the topological order of the nodes within the graph. Every node takes as input the output generated by all its predecessor nodes and generates its own output.

In the method by \citet{zhuge2024language}, edge selection within the graph is governed by a single parameter vector \(\theta\), where \(\theta_i\) represents the probability of sampling edge \(e_i\). This probability is optimized using \gls{RL} techniques, specifically the REINFORCE algorithm.
 The objective is to find the optimal \(\theta\) that maximizes the expected utility of the graph structures generated by this parameterization:
\begin{equation}
\label{eq:single_vector}
\theta^\star = \argmax_{\theta \in \Theta} \mathbb{E}_{G^\prime \sim D_{\theta}}[u_\mathcal{T}(G^\prime)],
\end{equation}
where \(\theta^\star\) represents the optimal parameter vector that maximizes the expected utility. Here, \(\Theta\) is the set of all possible parameter vectors, \(G^\prime\) denotes a graph structure sampled from the distribution \(D_{\theta}\) parametrized by \(\theta\), and \(u_\mathcal{T}(G^\prime)\) is the utility function that evaluates the performance of the graph \(G^\prime\).

\subsection{Input-Conditional Edge Probabilities}

Given the increasing complexity of tasks that language agents must handle, there is a need for these agents to exhibit a high degree of adaptability. This adaptability involves dynamically adjusting their computational routines based on the specific input, akin to the flexibility seen in Mixture of Experts (MoE) architectures, where different "experts" are selected based on the input to optimize processing \cite{jacobs1991adaptive}.

To enhance the adaptability and efficiency of language agents, we propose a novel method where edge probabilities are conditional on the input \(x\). Instead of using a fixed parameter vector, we introduce a function \(f\) that maps an input \(x\) to a vector of probabilities \(\theta\), tailored for that specific input:
\begin{equation}
\label{eq:conditional_prob}
f(x) = \theta, \quad x \sim D,
\end{equation}
where \(f\) is a function from the set of all possible functions \(\mathcal{F}\). The input \(x\) is sampled from the distribution \(D\), and \(\theta\) is the vector of probabilities generated by \(f\) for the input \(x\).

This approach allows the graph structure to dynamically adjust based on the input, improving the agent's ability to handle diverse tasks effectively. The optimization objective is then redefined to maximize the expected utility across different inputs and their corresponding graph structures:
\begin{equation}
\label{eq:optimization_objective}
f^\star = \argmax_{f \in \mathcal{F}} \mathbb{E}_{x \sim D}[\mathbb{E}_{G^\prime \sim D_{f(x)}}[\hat{u}(G^\prime(x))]],
\end{equation}
where \(f^\star\) is the optimal function that maps inputs to edge probabilities. Here, \(\mathcal{F}\) is the set of all possible functions, \(x\) is an input sampled from the distribution \(D\), \(G^\prime \sim D_{f(x)}\) represents a graph structure sampled from the distribution \(D_{f(x)}\) parametrized by the output of \(f(x)\), and \(\hat{u}(G^\prime(x))\) is a utility function that evaluates the output of the graph \(G^\prime\) executed on input \(x\).
This method is designed to be at least as effective as the previous approach, which directly updates edge probabilities. This is because the function can always be learned to be a constant function. We provide a proof of this in Appendix \ref{app:Theoretical Justification of Our Method}.
\begin{table*}[htbp]
\centering
\caption{Accuracy results for $k=5$ runs on the Crosswords Puzzle dataset by Dynamic Graph and Static Graph. We used LLama 3 8B instruction finetuned for LLM inference.}
\label{tab:crosswords}
\begin{tabular}{@{}lrrrrrrr@{}}
\toprule
\textbf{Method} & \multicolumn{5}{c}{Accuracy $x_i$ for Run $i$ (in \%)} & \textbf{Std} & \textbf{Mean} \\ 
\textbf{} & \textbf{} & \textbf{} & \textbf{} & \textbf{} & \textbf{} & \textbf{(in \%)} & \textbf{Acc. (in \%)} \\ 
\cmidrule(lr){2-6}
& $x_1$ & $x_2$ & $x_3$ & $x_4$ & $x_5$ & & \\ 
\midrule
Static Graph & 23.5 & 18.0 & 21.5 & 19.5 & 17.5 & 2.2 & 20.0 \\
Dynamic Graph & 21.5 & 18.5 & 19.5 & 21.5 & 21.0 & \textbf{1.2} & \textbf{20.4} \\
\bottomrule
\end{tabular}
\end{table*}
\subsection{Design of Our Methodology}

We employ the REINFORCE algorithm for training, providing a solid base for our initial experiments. Our goal is to learn the function \( f \), which dynamically adjusts graph structure using a pretrained \gls{LLM}.

Given the pivotal role of \glspl{LLM} in processing textual input, \( f \) must comprehend and reason about text to accurately assess task requirements and devise optimal strategies.

We use a learnable linear transformation layer to map the last hidden dimension of the \gls{LLM}'s output to a probability vector \( \theta \in \mathbb{R}^{|\mathcal{E}|} \) for edge selection in our graph-based model:
\(\theta = W \cdot h + b\), where \( W \in \mathbb{R}^{|\mathcal{E}| \times d} \) is the weight matrix, \( b \in \mathbb{R}^{|\mathcal{E}|} \) is the bias vector, and \( h \in \mathbb{R}^d \) is the output from the last hidden layer of the \gls{LLM}.

Following \cite{zhuge2024language}, we initialize the weights \( W \) to zero and set the bias \( b \) to reflect initial probabilities, guiding the initial learning phase.

However, initializing \( W \) with a normal distribution facilitates more effective gradient updates by breaking symmetry among neurons \cite{lecun2002efficient, glorot2010understanding}: \(W_{ij} \sim \mathcal{N}(0, \sigma^2)\). This diversity in initial weights allows neurons to learn different aspects of the input data, leading to more robust neural network models \cite{glorot2010understanding}.

\section{Experiments}
\begin{table*}[htbp]
\centering
\caption{Comparison of Dynamic Graph and Static Graph in terms of accuracy on the test sets. The expected number of edges refers to edges from the agents to the final decision node. We split the tasks into static tasks, which are reproduced from \cite{zhuge2024language}, and dynamic tasks, specifically designed to require language agents to change based on the input. The static tasks include accuracy results for the Crosswords Puzzle dataset and the Adversarial Agent experiment on the \gls{MMLU} dataset, while the dynamic tasks include accuracy with additional loss to reduce computational costs on the combined \gls{MMLU} and \gls{CMMLU} dataset. All accuracies are reported in percentages. Ratio refers to how much more likely it is to sample an edge from a truthful agent compared to an adversarial agent.}
\label{tab:method-comparison}
\begin{tabular}{@{}lrrrrrrr@{}}
\toprule
\textbf{Method} & \multicolumn{3}{c}{\textbf{Static Task}} & \multicolumn{4}{c}{\textbf{Dynamic Task}} \\ \cmidrule(lr){2-4} \cmidrule(lr){5-8}
 & Crosswords & \multicolumn{2}{c}{Adversarial Agents} & \multicolumn{2}{c}{Specialized Agents} & \multicolumn{2}{c}{+Edge Reduction} \\ \cmidrule(lr){2-2} \cmidrule(lr){3-4} \cmidrule(lr){5-6} \cmidrule(lr){7-8}
 & \textbf{Acc. (\%)} & \textbf{Acc. (\%)} & \textbf{Ratio} & \textbf{Acc. (\%)} & \textbf{Edges} & \textbf{Acc. (\%)} & \textbf{Edges} \\ \midrule
Static Graph & \hfill 20.0 & \hfill 44.6 & 1.52 &\hfill 44.7 & \hfill 5.01 & \hfill 42.1 & \hfill \textbf{2.60} \\
Dynamic Graph & \hfill \textbf{20.4} & \hfill \textbf{48.6} & \textbf{2.73} & \hfill \textbf{50.3} & \hfill \textbf{4.38} & \hfill \textbf{52.4} & \hfill 2.72 \\ \bottomrule
\end{tabular}
\end{table*}
\label{Experiments}
We replicated the experiments from \cite{zhuge2024language} to demonstrate that our method is at least as effective. Additionally, we introduced a new experimental framework to show that our method can significantly outperform the previous approach. In the following, we will refer to the previous method proposed by \citet{zhuge2024language} as Static Graph, and our method as Dynamic Graph. The code implementation we use is publicly available on GitHub\footnote{\url{https://github.com/lukasVierling/DynamicGPTSwarm}}, building upon the existing code base by \citet{zhuge2024language}\footnote{\url{https://github.com/metauto-ai/GPTSwarm}}.

\subsection{Crosswords Puzzle Experiment}
\label{Crosswords Puzzle Experiment}

In our initial experiment, we aimed to replicate the crosswords puzzle experiments to demonstrate that our method performs better than the previous approach. We utilized the instruction-finetuned version of \gls{LLM} known as LLama 3 8B \cite{meta_llama3_2024}, which has shown robust performance across various benchmarks despite its relatively smaller parameter size. This model was selected for its suitability for the crosswords puzzle task. For learning the edge probabilities (Eq. \ref{eq:optimization_objective}), we employed a Gemma-2B \gls{LLM} \cite{team2024gemma}, balancing model size and language understanding capabilities. The prompts used for this experiment are provided in Appendix \ref{app:Crosswords Prompt Set}.

Following \citet{zhuge2024language}, we explored two types of agents: the \gls{CoT} agent \cite{wei2022chain} and the Reflexion Agent \cite{shinn2024reflexion}. The \textbf{Chain of Thought} agent processes tasks in three logical steps: it initially analyzes the task in the first two steps and then outputs a solution based on the derived information in the final step. The \textbf{Reflexion Agent} initially solves the task with a greedy solution, receives feedback from an \gls{LLM}, and then outputs a new solution based on the feedback. The outputs of these agents are collected by a \textbf{Return All} agent, which returns all solutions. There was a third agent, the \textbf{Tree of Thought} agent \cite{yao2024tree}, but it was excluded due to the quadratic scaling of potential edges with the increase in nodes within a graph, in order to maintain computational feasibility.

We used a mini crosswords dataset of 156 5x5 puzzles from GooBix \cite{yao2024tree} and evaluated performance based on correct words for direct puzzle-solving assessment. Details of the dataset and language agent's graph structure, along with training parameters, are provided in Appendix \ref{app: Mini Crosswords Puzzle} and \ref{app:Crosswords Puzzle Experiment}.

\begin{figure*}[h]
    \centering
    \begin{subfigure}[t]{0.49\textwidth}
        \centering
        \includegraphics[width=0.85\linewidth]{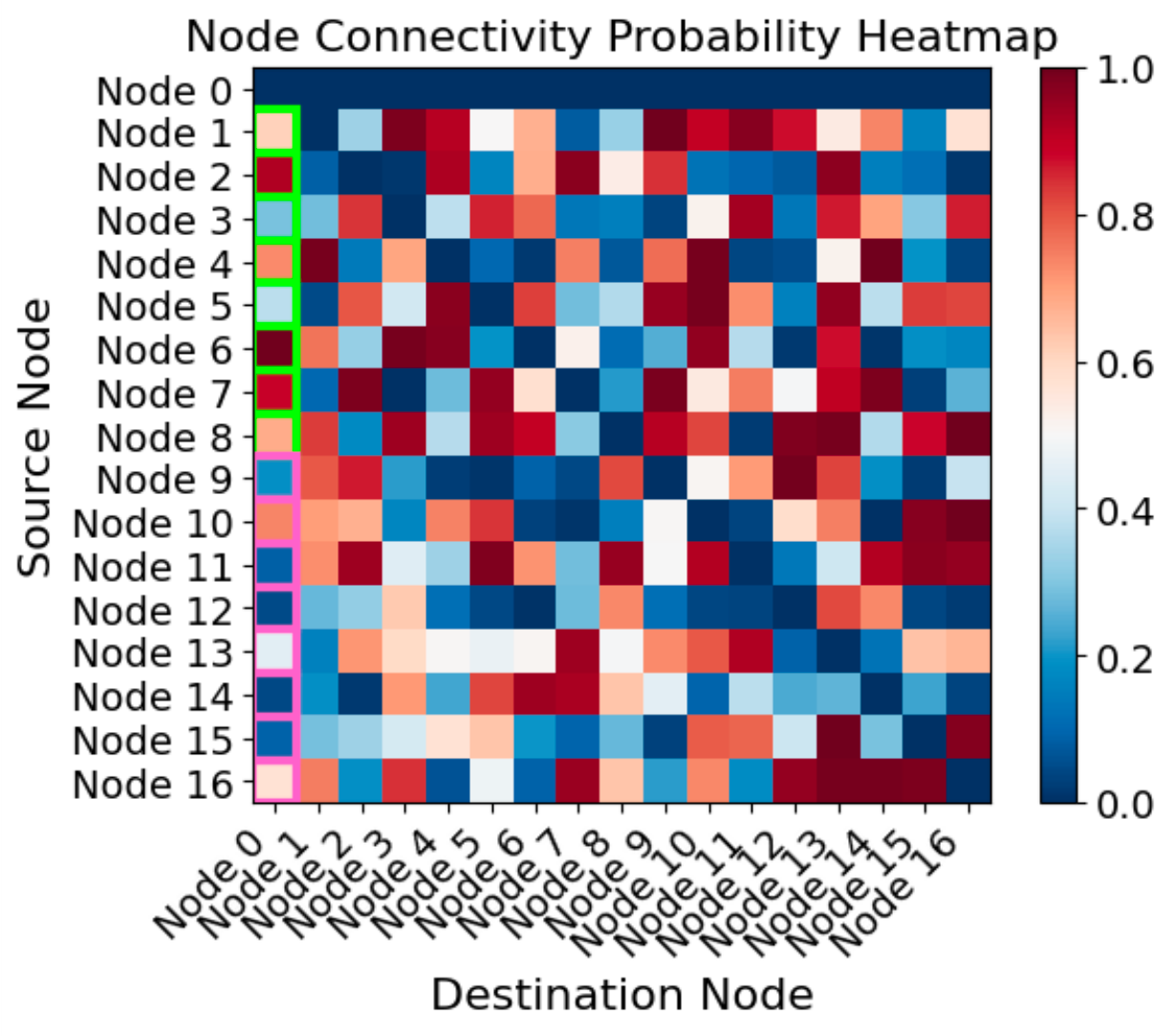}
        \caption{Probabilities for sampling an edge in the graph learned by Dynamic Graph.}
        \label{fig:heatmap_edge}
    \end{subfigure}%
    \hfill
    \begin{subfigure}[t]{0.49\textwidth}
        \centering
        \includegraphics[width=0.85\linewidth]{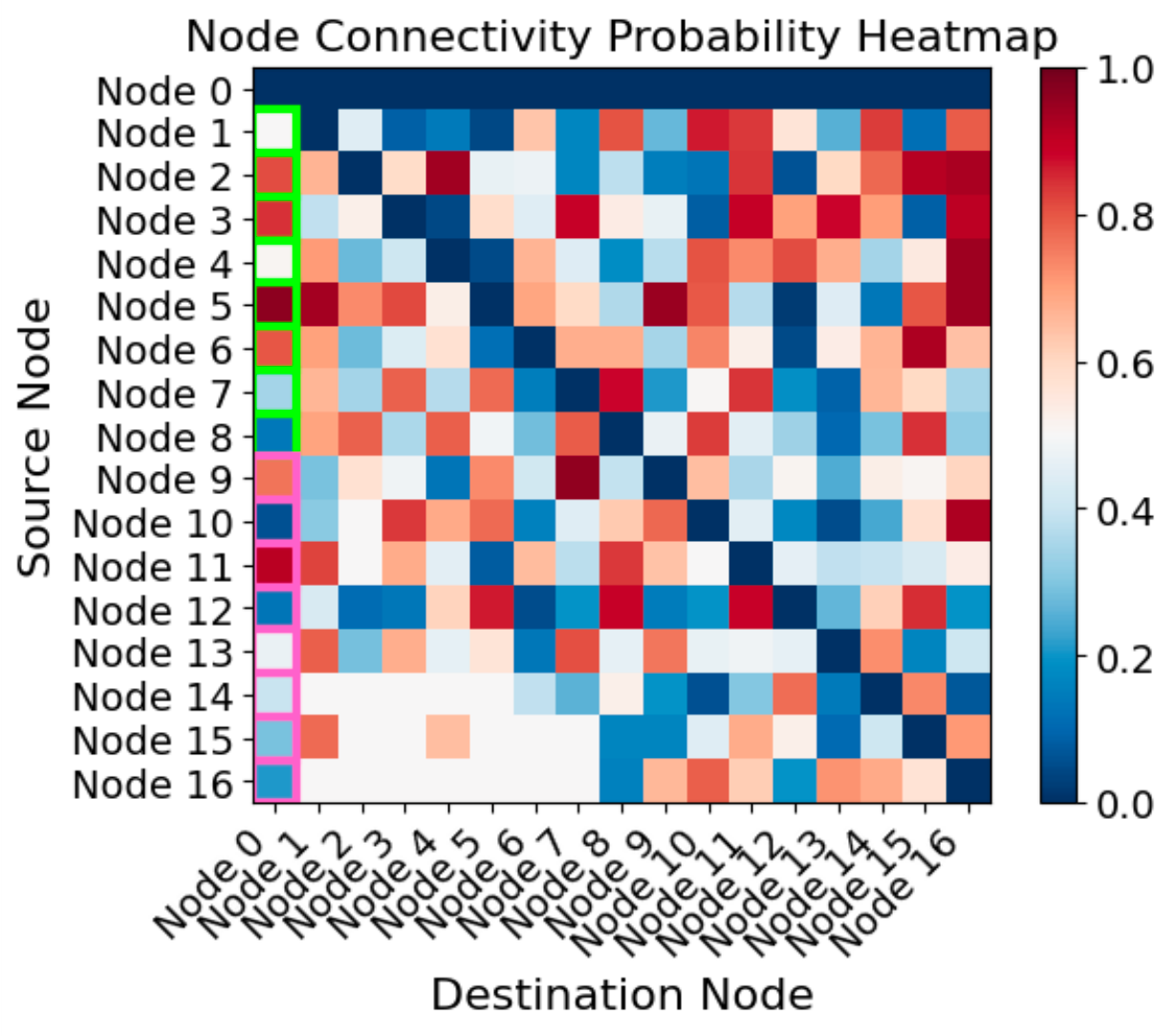}
        \caption{Probabilities for sampling an edge in the graph learned by Static Graph.}
        \label{fig:heatmap_noedge}
    \end{subfigure}
    \caption{Comparative visualization of edge probabilities in graphs learned by Dynamic Graph and Static Graph. Node 0 is the final decision node, nodes 1 to 8 are truthful agents, and nodes 9 to 16 are adversarial agents. Self-loops and connections from the final decision node to any other node are set to 0.}
    \label{fig:combined_heatmaps}
\end{figure*}

\textbf{Results and Analysis:}
Due to the inherent randomness in the graph sampling process during evaluation, we tested our method on the test set five times. For the Static Graph approach we sampled five graphs. The results, detailed in Table \ref{tab:crosswords}, demonstrate that Dynamic Graph not only achieves higher average accuracy but also exhibits less variance across these runs, indicating enhanced performance and consistency from language agents using our method.

\subsection{Adversarial Agent Detection}
\label{Adversarial Agent Detection}
We replicated experiments from \citet{zhuge2024language} with the \gls{MMLU} benchmark to validate our method's ability to identify adversarial agents and compare it with Static Graph.

\textbf{Experimental Setup:}
We used the instruction-finetuned Gemma-7B model, configuring seventeen agents: eight adversarial, eight truthful, and one final decision agent. Adversarial agents output lies, while truthful agents provide honest answers. The final decision agent performs majority voting on the inputs. Smaller models like Gemma-7B struggled with untruthful responses, frequently defaulting to truthful answers. To address this, adversarial agents were modified to consistently respond with "A".

The dataset used was the \gls{MMLU} dataset \cite{hendrycks2021measuring}, which includes multiple-choice questions with four answer options, covering 57 topics. This dataset is a standard for assessing the world knowledge and problem-solving abilities of \glspl{LLM}. Samples of this dataset are in Appendix \ref{app:MMLU Dataset}.

Following the original experiment's configuration, we conducted 200 iterations with a batch size of 4, a learning rate of 0.0001, and used the Adam optimizer. Both methods were trained on the \gls{MMLU} dev set and tested on 1000 questions from the val set. Prompts used for this experiment are in Appendix \ref{app:MMLU Prompt Set}.

\textbf{Results and Analysis}

The Dynamic Graph effectively identified and excluded most adversarial agents. The average probability across the test set was calculated as
\[
\bar{\theta} = \frac{1}{|D|} \sum_{x \in D} f(x),
\]
where $\bar{\theta}$ is the average probability, $D$ is the test set, and $f(x)$ is the function outputting edge probabilities.

Figures \ref{fig:heatmap_edge} and \ref{fig:heatmap_noedge} show heatmaps of edge probabilities in the graph. Our method demonstrated higher effectiveness in identifying adversarial agents, with probabilities for edges from truthful agents close to 1 and from adversarial agents near 0.

Critical edges are those from agents (nodes 1 to 16) to the final decision agent (node 0). The heatmaps illustrate that our method assigns high probabilities to edges from truthful agents and low probabilities to those from adversarial agents, enhancing robustness and accuracy. Outliers at nodes 5 or 10 had minimal impact due to the majority vote mechanism. This is evident from the ratio between the probability of sampling a critical edge from truthful agents compared to adversarial agents, which increased from 1.52 in the Static Graph to 2.73 in the Dynamic Graph.

Overall, the Dynamic Graph improves the detection and exclusion of adversarial agents, leading to higher accuracy, as shown in Table \ref{tab:method-comparison}.
\begin{figure*}[h]
    \centering
    \begin{subfigure}[t]{0.32\textwidth}
        \centering
        \includegraphics[width=\linewidth]{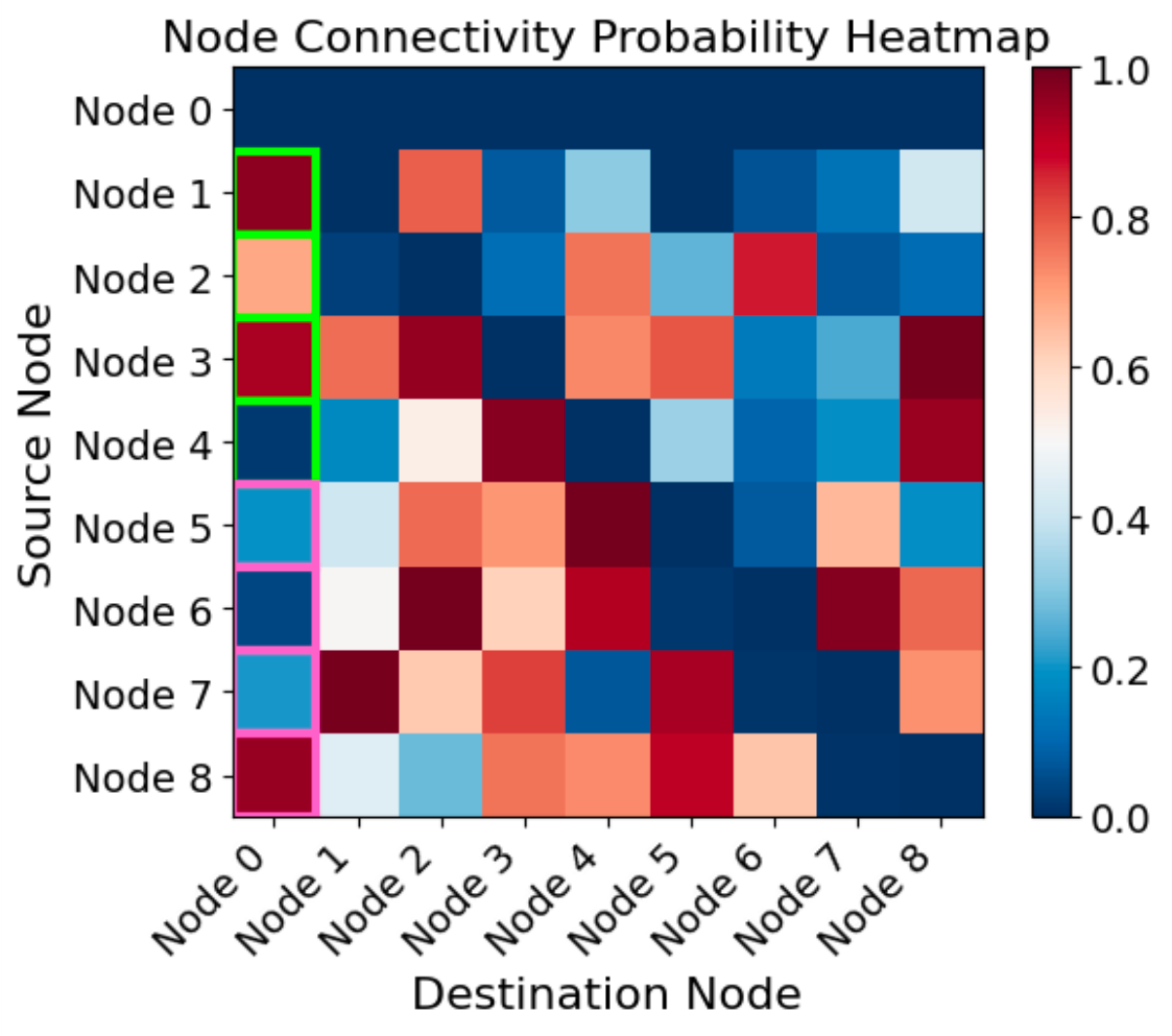}
        \caption{Probabilities for sampling an edge on the \gls{MMLU} dataset in the graph learned by Dynamic Graph.}
        \label{fig:heatmap_edge_specialized_mmlu}
    \end{subfigure}%
    \hfill
    \begin{subfigure}[t]{0.32\textwidth}
        \centering
        \includegraphics[width=\linewidth]{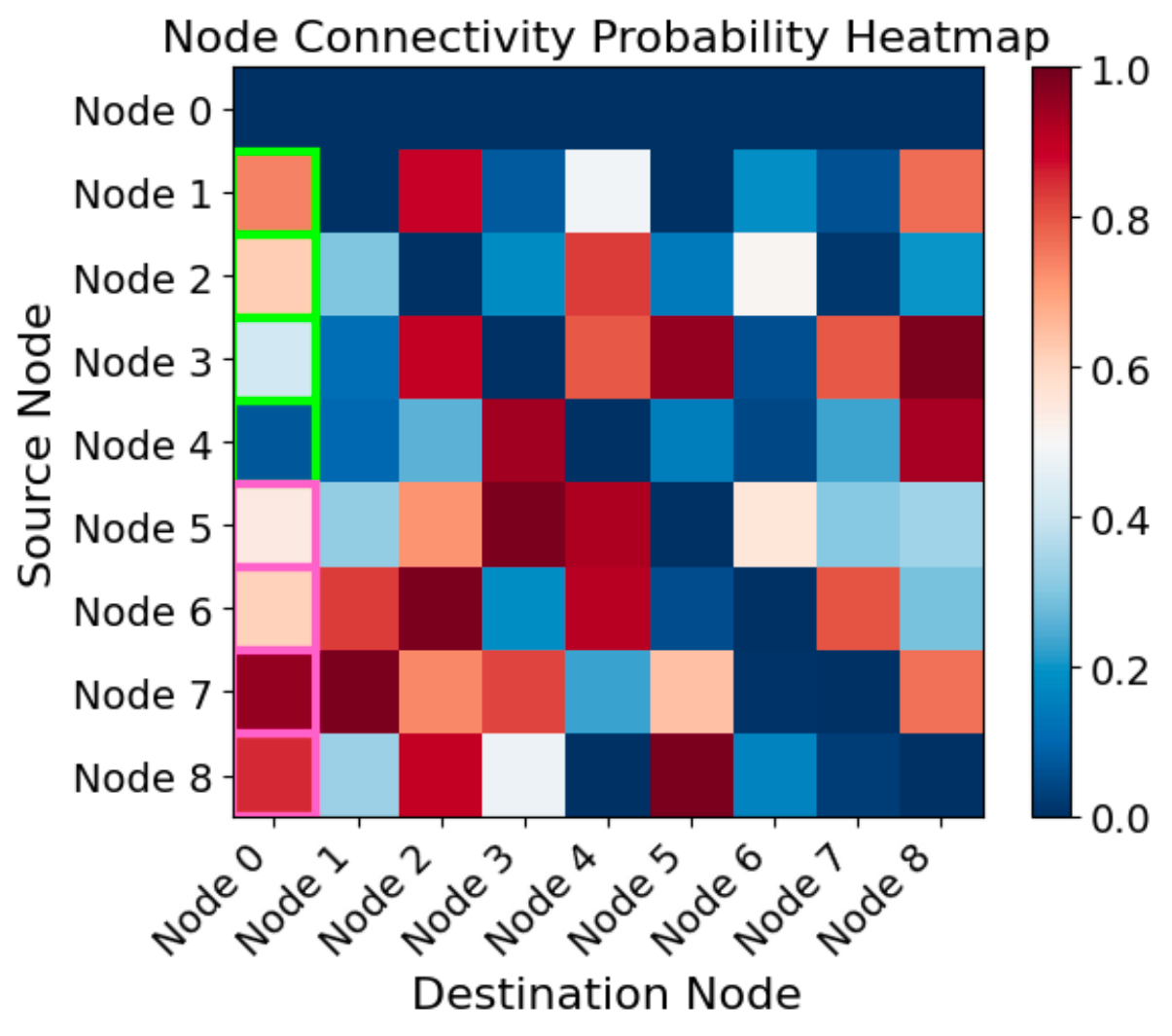}
        \caption{Probabilities for sampling an edge on the \gls{CMMLU} dataset in the graph learned by Dynamic Graph.}
        \label{fig:heatmap_edge_specialized_cmmlu}
    \end{subfigure}%
    \hfill
    \begin{subfigure}[t]{0.32\textwidth}
        \centering
        \includegraphics[width=\linewidth]{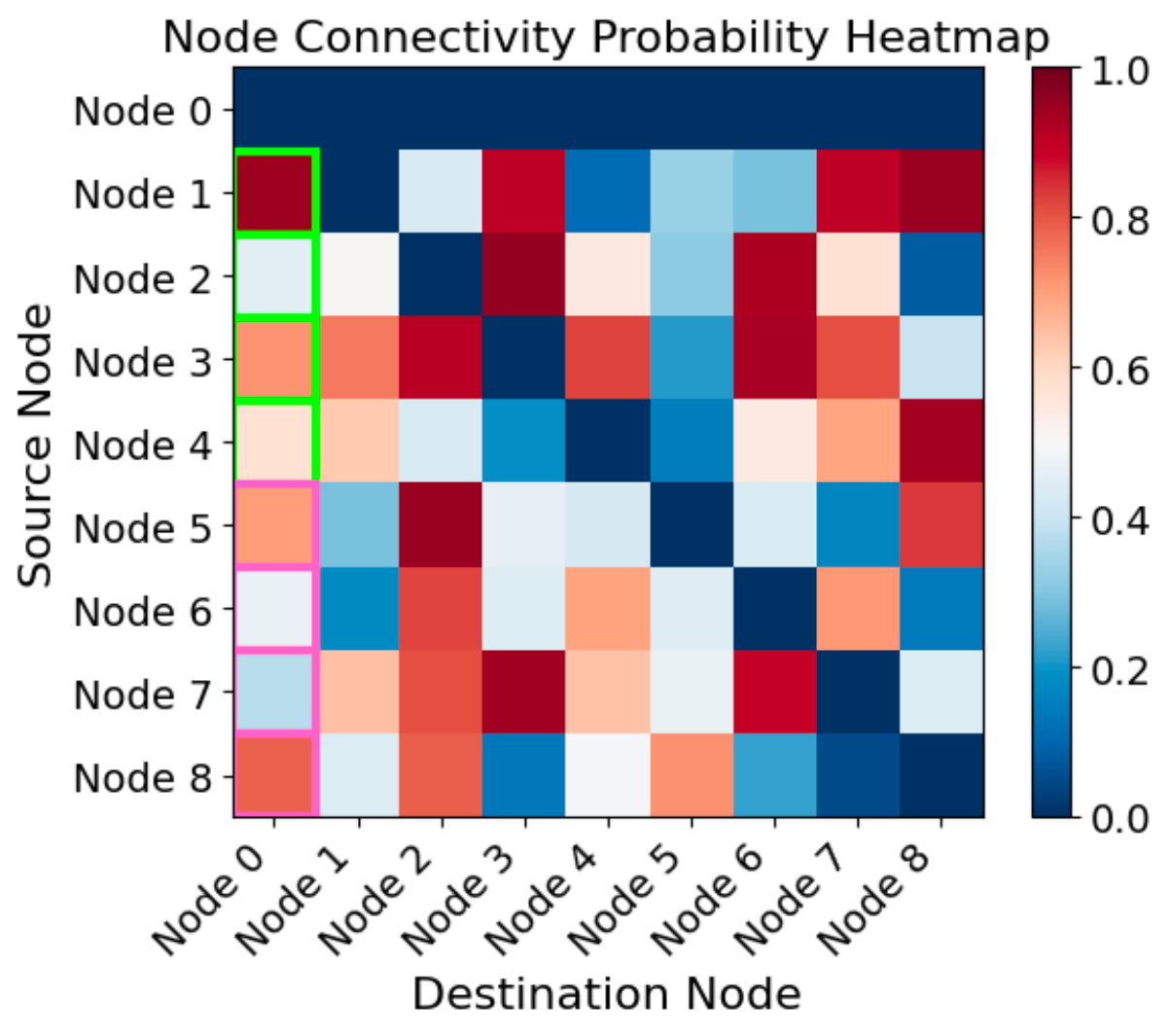}
        \caption{Probabilities for sampling an edge in the graph by Static Graph.}
        \label{fig:heatmap_no_edge_specialized}
    \end{subfigure}
    \caption{Comparative visualization of edge probabilities on MMLU and CMMLU datasets. Node 0 is the final decision node, nodes 1 to 4 are truthful agents using Gemma-7B-it, and nodes 5 to 8 are truthful agents using BlueLM-7B-chat. Notably, self-loops as well as connections from the final decision node to any other node are not allowed and thereby 0.}
    \label{fig:combined_heatmaps_specialized}
\end{figure*}
\subsection{Specialized Agents Experiment}
\label{Specialized Agents Experiment}
\label{specializedAgents}

This experiment aimed to demonstrate the superiority of our method over the Static Graph approach. We trained a language agent on a diverse dataset requiring input-specific adaptation. Dynamic Graph adjusts to the specific characteristics of the input, unlike the Static Graph method, which aims for a generalized solution, often resulting in decreased performance.

\textbf{Experimental Setup:}
We configured eight truthful agents and a final decision agent, evenly divided between two language models: four based on Gemma-7B-it \cite{team2024gemma} and four on BlueLM-7B-Chat \cite{2023bluelm}. Despite their similar sizes, these \glspl{LLM} show varying performance (Appenidx \ref{app:Specialized Agents Experiment}) depending on the dataset used. Gemma-7B-it excels with the \gls{MMLU} dataset \cite{hendrycks2021measuring}, while BlueLM-7B-Chat performs better on the \gls{CMMLU} dataset \cite{li2023cmmlu}.

\textbf{Results and Analysis:}
Our experiment demonstrated that Dynamic Graph effectively identified performance differences between the two \glspl{LLM} and assigned higher probabilities to agents better suited for specific questions. Visualizations of adjacency matrices showed that the Static Graph found a more general solution, incorporating almost all agents with high probabilities (Figure \ref{fig:heatmap_no_edge_specialized}).

Heatmaps for the \gls{CMMLU} (Figure \ref{fig:heatmap_edge_specialized_cmmlu}) and \gls{MMLU} (Figure \ref{fig:heatmap_edge_specialized_mmlu}) datasets displayed the average probability of sampling edges. Edges from Gemma-7B-it agents are highlighted in green, and those from BlueLM-7B-chat agents in pink. The model consistently assigned higher probabilities to the \glspl{LLM} that performed better on these datasets, aiming for higher accuracy. Detailed edge probabilities are listed in Table \ref{app:Specialized Agents Experiment}.

Our approach selected relevant agents based on input, reducing computational load by minimizing the number of agents needed for final decisions, thus saving resources. Table \ref{tab:method-comparison} shows that our method improved accuracy by nearly 6\% over Static Graph on the test set.

These results highlight the effectiveness of task-dependent graph construction for edge optimization. By dynamically adapting the graph and leveraging agent specifications, Dynamic Graph outperforms Static Graph, improving the performance of language agents on diverse tasks.

\subsection{Edge Reduction on Specialized Agents}
\label{Edge Reduction on Specialized Agents}
In this graph framework, reducing the number of edges can lower computational costs. We introduced an additional loss function during training to prioritize key nodes and reduce unnecessary internal communications. The loss function used was: \(
L(\theta) = \delta \cdot \sum_{i=1}^{|\mathcal{E}|} |\theta_i|
\)
with \(\delta\) set to 0.1. This sparsity-inducing loss \cite{tibshirani1996regression} was applied using backpropagation to guide the model toward fewer edges, enhancing computational efficiency. The experimental setup mirrored Section \ref{Specialized Agents Experiment}, maintaining identical agent and training parameters.

\textbf{Results and Analysis:}
\begin{figure*}[h]
    \centering
    \begin{subfigure}[t]{0.32\textwidth}
        \centering
        \includegraphics[width=\linewidth]{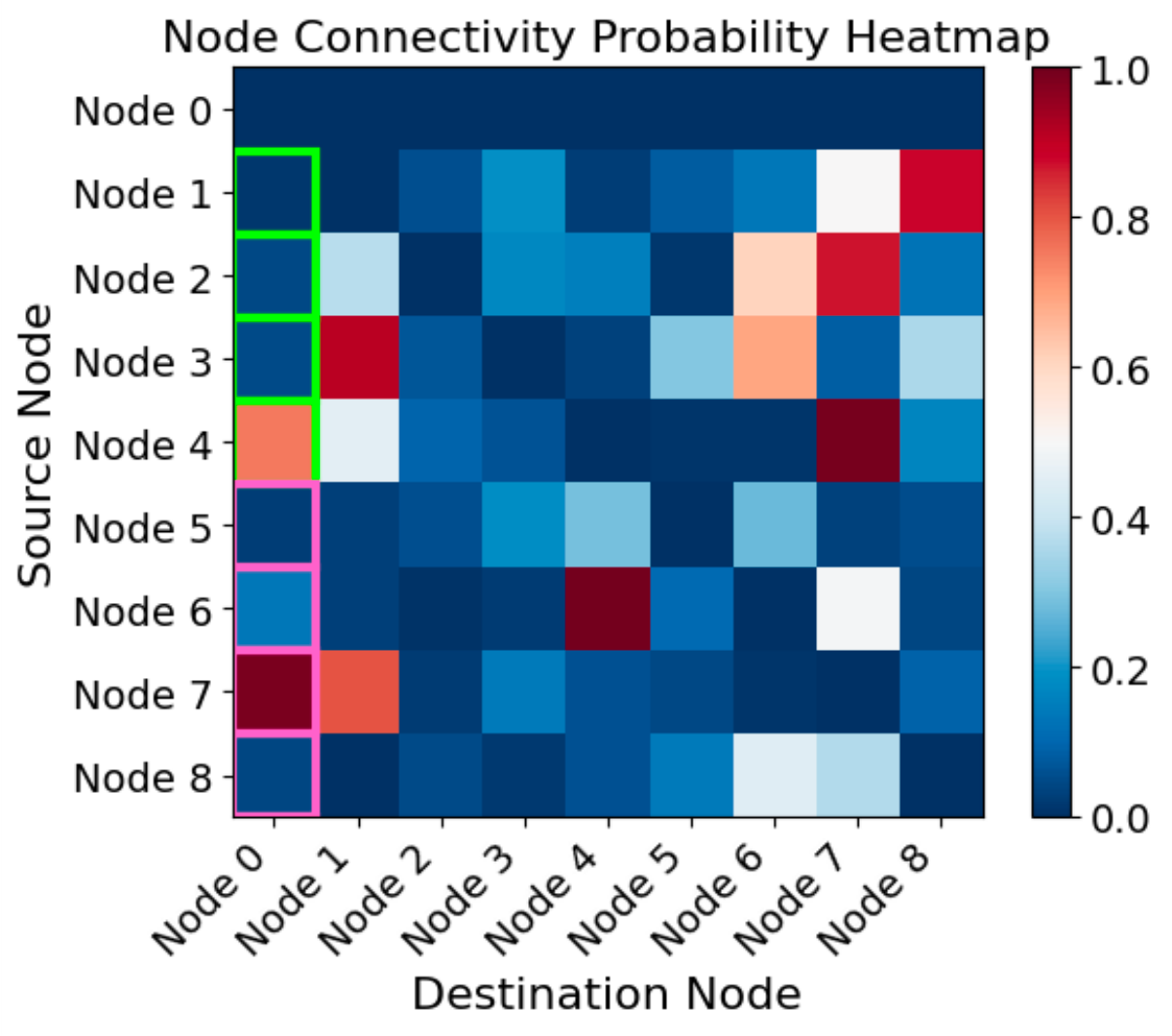}
        \caption{Edge probabilities on \gls{MMLU} with Dynamic Graph and additional loss.}
        \label{fig:reduce_edge_old}
    \end{subfigure}%
    \hfill
    \begin{subfigure}[t]{0.32\textwidth}
        \centering
        \includegraphics[width=\linewidth]{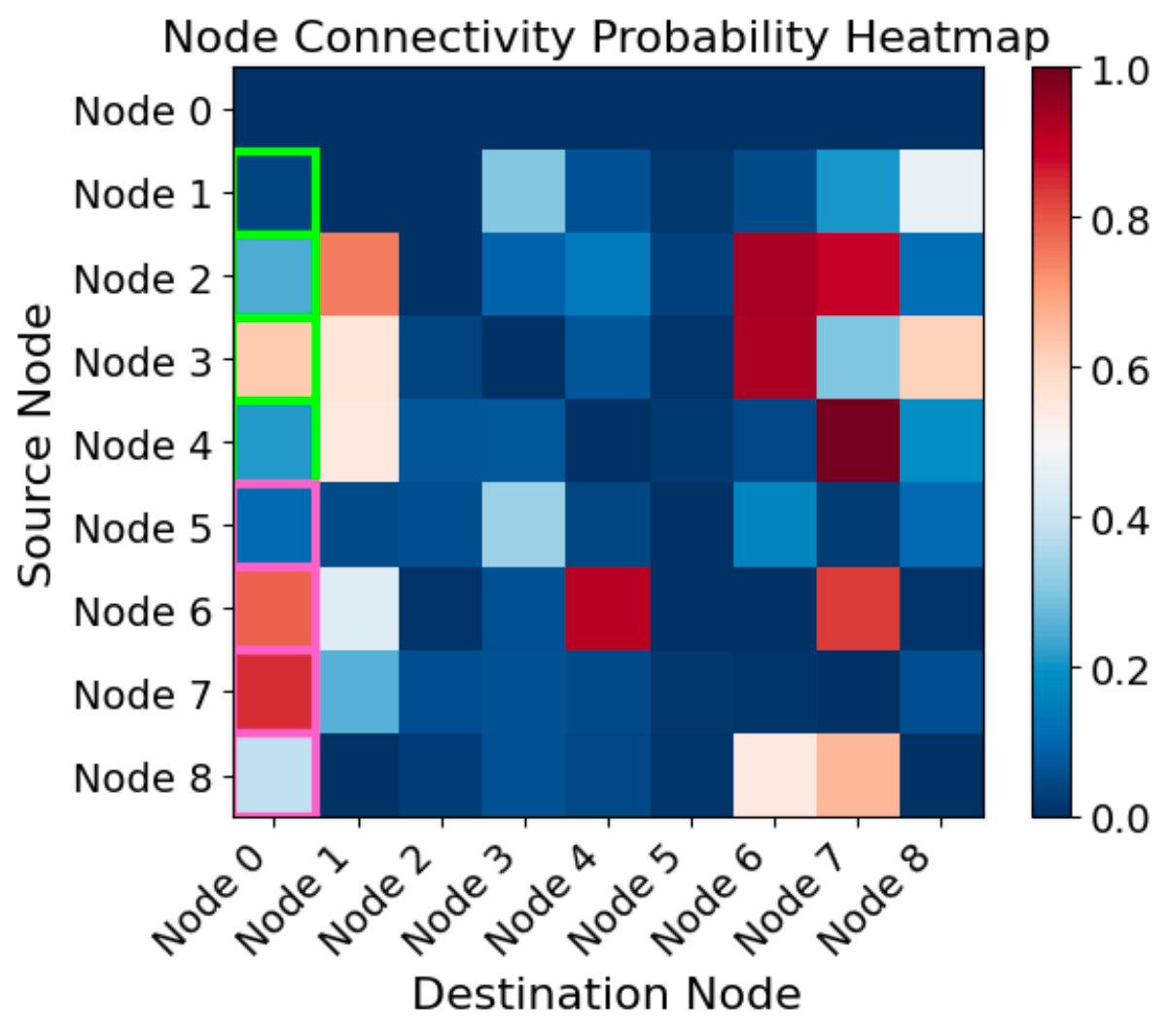}
        \caption{Edge probabilities on \gls{CMMLU} with Dynamic Graph and additional loss.}
        \label{fig:reduce_edge_mmlu}
    \end{subfigure}%
    \hfill
    \begin{subfigure}[t]{0.32\textwidth}
        \centering
        \includegraphics[width=\linewidth]{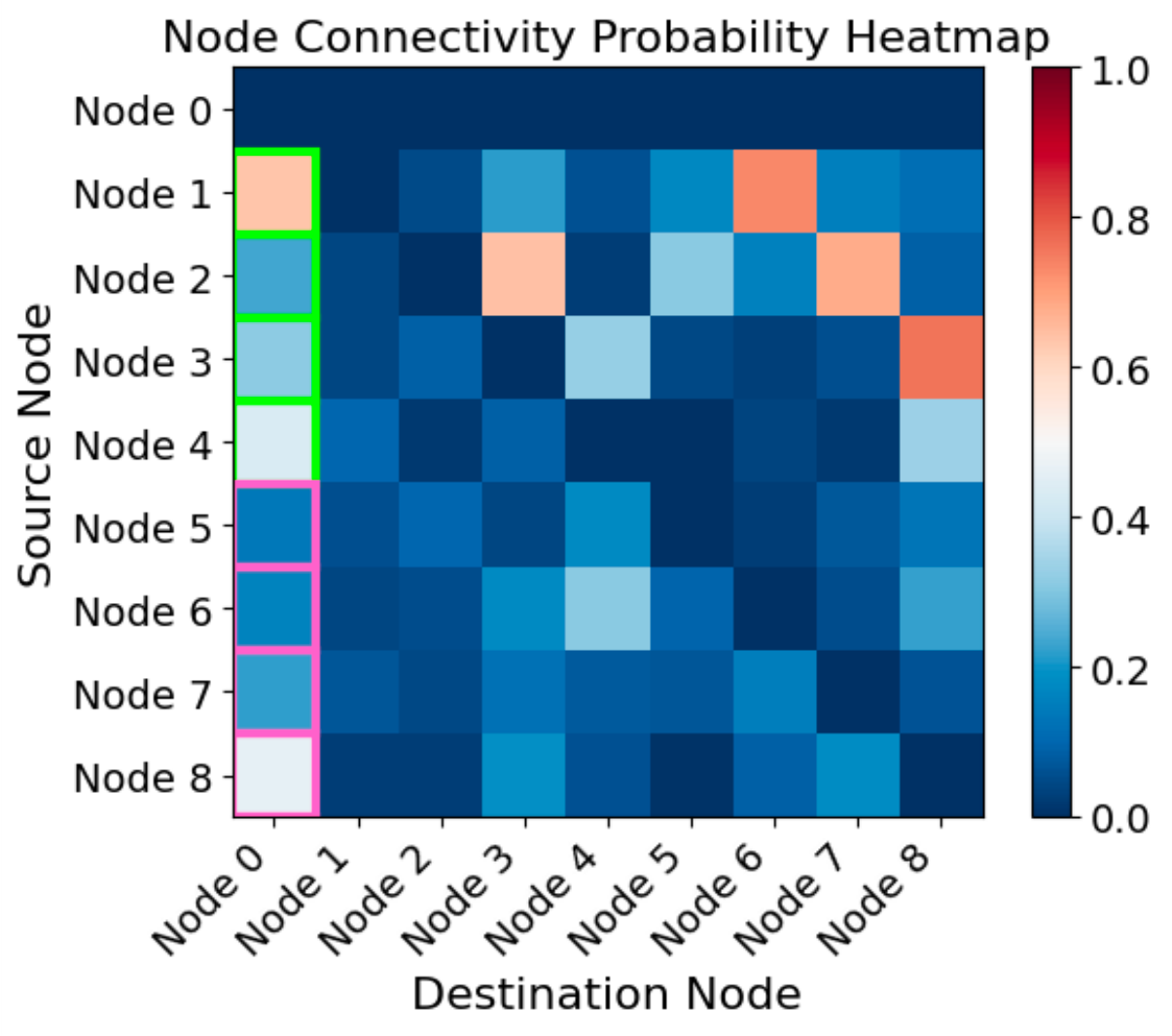}
        \caption{Edge probabilities with Static Graph and additional loss.}
        \label{fig:reduce_edge_cmmlu}
    \end{subfigure}
    \caption{Comparative visualization of edge probabilities learned by different methods on old and MMLU datasets. Node 0 is the final decision node, nodes 1 to 4 are Gemma-7B-it agents, and nodes 5 to 8 are BlueLM-7B-chat agents. Self-loops and connections from the final decision node to any other node are not allowed.}
    \label{fig:combined_reduce_edge}
\end{figure*}
Dynamic Graph showed enhanced performance over Static Graph. While the earlier method's performance declined, our approach improved, indicating that edge reduction helps identify relevant nodes and generate input-dependent graphs effectively.

Figure \ref{fig:reduce_edge_old} shows edge probabilities with Static Graph, while Figures \ref{fig:reduce_edge_mmlu} and \ref{fig:reduce_edge_cmmlu} show edge probabilities with Dynamic Graph on the \gls{MMLU} and \gls{CMMLU} datasets. Dynamic Graph adapts preferences based on input, especially noticeable in the \gls{CMMLU} dataset. Table \ref{tab:method-comparison} shows our method outperformed the previous one by over 10\%, identifying relevant agents and demonstrating robustness to changes in loss functions. It should be noted that Dynamic Graph sampled slightly more edges from truthful agents to the final decision agents.

\subsection{Ablation Study}
\label{Ablation Study}
We investigated the role of the \gls{LLM} in predicting edge probabilities. We compared our method, which uses transformer layers for textual representation, with a baseline model using a pretrained embedding matrix from Gemma 2B. The baseline averages embeddings and maps them to edge probabilities via a linear layer. This aligns with prior studies suggesting that averaging embeddings can capture essential textual information \cite{mikolov2013distributed,arora2017simple}. The experiment setup was the same as in Section \ref{Specialized Agents Experiment}.

\textbf{Results and Analysis:}
The baseline method struggled to differentiate based on input text, often defaulting to a constant vector output. Despite this, it performed better than earlier methods, achieving an accuracy of 45.7\%. The heatmap of probabilities is in Appendix \ref{app:Heatmap Ablation Study}.

These results highlight the importance of a component capable of comprehending language and text to generate task-dependent graphs accurately.

\section{Conclusion}
\label{Conclusion}
This work presents a new approach for edge optimization in graph frameworks for language agents, as introduced by \citet{zhuge2024language}. Unlike Static Graph, which used a single vector of probabilities for edge sampling, Dynamic Graph learns a function \( f \) that dynamically maps the agent's input to edge probabilities. This generalizes the previous approach, which is a special case where \( f \) is constant. We train this function using a neural network built on a pretrained \gls{LLM}. 

Experimental results in Section \ref{Experiments} show Dynamic Graph consistently outperforms Static Graph across all tasks. Specifically, in a task-dependent graph construction experiment (Section \ref{specializedAgents}), Dynamic Graph exceeded Static Graph by nearly 6\% accuracy. This flexibility allows language agents to adjust their communication strategies based on input, processing input more effectively and enhancing their adaptability across various tasks.

\section{Limitations}
\label{Future Work and Limitations}

Our work establishes an experimental foundation demonstrating the potential benefits of dynamically adjusting language agents based on their input. However, further research is necessary to explore these concepts on a larger scale. 

To deepen understanding, the effect of dynamic language agents should be investigated using agent swarms comprising a greater number of agents and employing larger \glspl{LLM}. Additionally, while our research aimed to demonstrate the comparative efficacy of our method against that proposed by \citet{zhuge2024language}, with superiority demonstrated through an additional experiment utilizing a mixed dataset, our experiments were confined to those conducted by \citet{zhuge2024language} employing edge optimization. While this comparison suffices to showcase the superior performance of our method in those specific experiments, future research should assess our method's capabilities across a broader spectrum of datasets and tasks, such as code generation.

Furthermore, exploring input-dependent edge generation could empower agents to dynamically adjust graph complexity based on input difficulty. For instance, a language agent might utilize a subgraph induced by nodes \( V' \subseteq V \) to process an input \( x \) efficiently, thereby conserving computational resources.

Moreover, we propose investigating dynamic node generation, which would enable agents to generate, add, or remove nodes and edges during execution. This capability could enhance flexibility and empower graphs to effectively handle diverse inputs.

\section{Potential Risks}

Language Agents can substantially extend the capabilities of \glspl{LLM}, allowing them to interact with their environment through a multitude of ways. Consequently, there is a concern that these advancements could lead to widespread automation, potentially displacing human labor on a large scale \citep{brynjolfsson2015will}. Moreover, there is a risk that these language agents could be exploited for illegal or dangerous activities \citep{brundage2018malicious}. 

We acknowledge the potential for our research, contributing to this field, to have harmful effects on society. Therefore, we advocate for more work on AI safety measures and the controlled deployment of AI technologies \citep{amodei2016concrete}.

\bibliography{acl_latex}

\appendix

\section{Theoretical Justification of Our Method}
\label{app:Theoretical Justification of Our Method}

Our approach is naturally designed to be as effective as or better than the previous approach, particularly when learning a constant function defined as:
\begin{equation}
\label{eq:constant_function}
f(x) = \theta^\star \in \mathbb{R}^{|\mathcal{E}|}
\end{equation}
where \( \theta^\star \) is derived from Equation (1), ensuring that the performance matches or exceeds that of the existing methodology.

The original optimization goal set forth by \citet{zhuge2024language} is:
\begin{equation}
\label{eq:original_goal}
\theta^\star = \argmax_{\theta \in \Theta} \mathbb{E}_{G^\prime \sim D_{\theta}}[u_\mathcal{T}(G^\prime)]
\end{equation}

Considering the utility function as the average utility over the current batch, we can rewrite this as:
\begin{equation}
\label{eq:average_utility}
\theta^\star = \argmax_{\theta \in \Theta} \mathbb{E}_{G^\prime \sim D_{\theta}} \left[\frac{1}{B} \sum_{i=1}^{B} \hat{u}(G^\prime(x))\right]
\end{equation}

Assuming a sufficiently large batch size, this average utility represents an unbiased estimator of the expected utility, allowing us to reframe it as:
\begin{equation}
\label{eq:unbiased_estimator}
\theta^\star = \argmax_{\theta \in \Theta} \mathbb{E}_{G^\prime \sim D_{\theta}}[\mathbb{E}_{x \sim D}[\hat{u}(G^\prime(x))]]
\end{equation}

Using the commutativity of expected values, it simplifies to:
\begin{equation}
\label{eq:commutativity}
\theta^\star = \argmax_{\theta \in \Theta} \mathbb{E}_{x \sim D}[\mathbb{E}_{G^\prime \sim D_{\theta}}[\hat{u}(G^\prime(x))]]
\end{equation}

Defining \( \theta^\star \) as the solution to this optimization.

By constraining the function set \( \mathcal{F} \) to constant functions, $\mathcal{F}_c$, we align our new optimization goal with the original objective:
\begin{equation}
\label{eq:aligned_goal}
\begin{aligned}
\argmax_{f \in \mathcal{F}_c} \mathbb{E}_{x \sim D}[\mathbb{E}_{G^\prime \sim D_{f(x)}}[\hat{u}(G^\prime(x))]] \\ = \argmax_{\theta \in \Theta} \mathbb{E}_{x \sim D}[\mathbb{E}_{G^\prime \sim D_{\theta}}[\hat{u}(G^\prime(x))]] \\
= \theta^\star
\end{aligned}
\end{equation}

Since the set of constant functions \(\mathcal{F}_c \subseteq \mathcal{F}\), our solution is at least as good as the solution found by the method introduced by \citet{zhuge2024language}.

\section{Experiments}
\subsection{Crosswords Puzzle Experiment}
\label{app:Crosswords Puzzle Experiment}
\begin{figure}[h]
    \centering
    \includegraphics[width=0.9\linewidth]{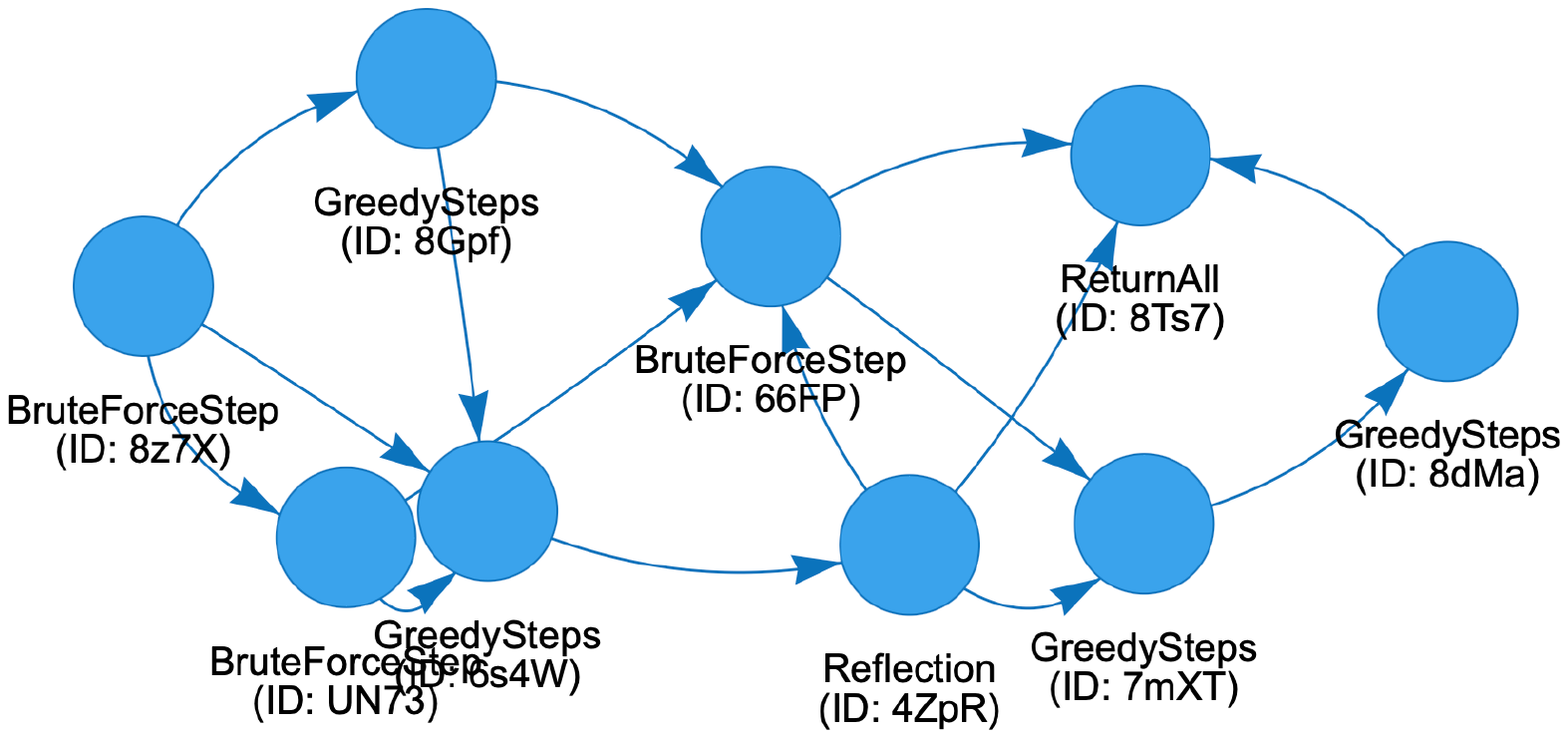}
    \caption{This is a sample graph for the crosswords experiment setup.}
    \label{fig:crosswords_graph}
\end{figure}

We provide an example of the language agent graph of the Crosswords Puzzle experiment in Figure \ref{fig:crosswords_graph}.

We used a mini crosswords dataset comprising 156 5x5 crossword puzzles collected from GooBix\footnote{\url{https://www.goobix.com/crosswords/}} as described in \cite{yao2024tree}. Performance on this dataset was evaluated using three primary metrics: correct letters, correct words, and correct games. Consistent with previous studies, our evaluation focused on the number of correct words, allowing for a direct assessment of puzzle-solving effectiveness based on the provided clues. Samples of this dataset are available in Appendix \ref{app: Mini Crosswords Puzzle}.

We used the same subset of 20 crossword puzzles for training and testing as \cite{zhuge2024language,yao2024tree}. Our approach, with a greater number of learnable parameters and a reduced learning rate, required extended training iterations. We increased the iteration count from 10 to 40 and decreased the batch size from 20 to 5, ensuring a total of 200 examples were presented during training. The initial edge-sampling probability was set at 0.1, with a learning rate of 0.0001 using the Adam optimizer. We reported the best state word accuracy \cite{yao2024tree,zhuge2024language}, indicating the accuracy of the best-proposed solution.
\subsection{Specialized Agents Experiment}
\label{app:Specialized Agents Experiment}

\textbf{Training Details:}
We used a combined dataset from \gls{MMLU} and \gls{CMMLU}. Each dataset contains multiple-choice questions, testing agents on a wide range of domains and languages. The \gls{CMMLU} benchmark, with questions in Mandarin Chinese, assesses \glspl{LLM} across 67 topics and includes linguistic and culturally specific content (Appendix \ref{app:CMMLU Dataset}).

Training was conducted over 200 iterations with a batch size of 4 and a learning rate of 0.0001 using the Adam optimizer. We trained on the dev sets of \gls{MMLU} and \gls{CMMLU}, and tested on 1000 questions from the \gls{MMLU} validation set and the \gls{CMMLU} test set.

We provide the performance of both \glspl{LLM} used for Section \ref{Specialized Agents Experiment} on the \gls{MMLU} and \gls{CMMLU} benchmarks. Since we couldn't find a reported value for the performance of Gemma on the \gls{CMMLU} dataset, we evaluated the model ourselves in Table \ref{tab:cmmlu_mmlu}.

\begin{table}[h]
\centering
\caption{Comparison of \gls{LLM} performance on the MMLU and CMMLU benchmarks.}
\label{tab:cmmlu_mmlu}
\begin{tabular}{@{}lcc@{}}
\toprule
\textbf{Model Name} & \multicolumn{2}{c}{\textbf{Accuracy (\%)}} \\
\cmidrule(lr){2-3}
& \textbf{CMMLU} & \textbf{MMLU} \\
\midrule
BlueLM-7B-Chat & \textbf{72.7} & 50.7 \\
Gemma-7B & 37.0 & \textbf{64.3} \\
\bottomrule
\end{tabular}
\end{table}

\textbf{Results:}
We provide a detailed breakdown of the relevant edge probabilities from agents to the final decision node in Table \ref{tab:prob_comparison}. The table shows that our method was able to detect the differences in the \glspl{LLM}' abilities and adjust their contribution to the final result based on the input's origin dataset.
\begin{table*}[h]
\centering
\caption{Comparison of Dynamic Graph with Static Graph method. We report the probabilities for sampling edges from the agents to the final decision node. For our method we further report the average probability over the test set of both the MMLU and CMMLU datasets.}
\label{tab:prob_comparison}
\begin{tabular}{@{}crrrr@{}}
\toprule
\textbf{Node} & \multicolumn{3}{c}{\textbf{Dynamic Graph}} & \textbf{Static Graph} \\
\cmidrule(lr){2-4}
& \textbf{CMMLU} & \textbf{MMLU} & \textbf{Difference} & \\
\midrule
\multicolumn{5}{c}{\textbf{Gemma-7B-It}} \\
\midrule
1 & 0.792 & 0.974 & \down{-0.182} & 0.942 \\
2 & 0.629 & 0.716 & \down{-0.087} & 0.456 \\
3 & 0.460 & 0.936 & \down{-0.476} & 0.717 \\
4 & 0.059 & 0.023 & \up{0.036} & 0.567 \\
\midrule
\multicolumn{5}{c}{\textbf{BlueLM-7B-Chat}} \\
\midrule
5 & 0.519 & 0.211 & \up{0.308} & 0.701 \\
6 & 0.556 & 0.063 & \up{0.493} & 0.471 \\
7 & 0.938 & 0.075 & \up{0.863} & 0.374 \\
8 & 0.852 & 0.956 & \down{-0.104} & 0.782 \\
\bottomrule
\end{tabular}
\end{table*}
\subsection{Ablation Study}
\label{app:Heatmap Ablation Study}
In this appendix, we provide the heatmap depicting the edge probabilities for the ablation study described in \ref{Ablation Study}. The heatmap visualizes the probabilities assigned to different edges in the graph based on the input text.

\begin{figure}[H]
    \centering
    \includegraphics[width=0.7\linewidth]{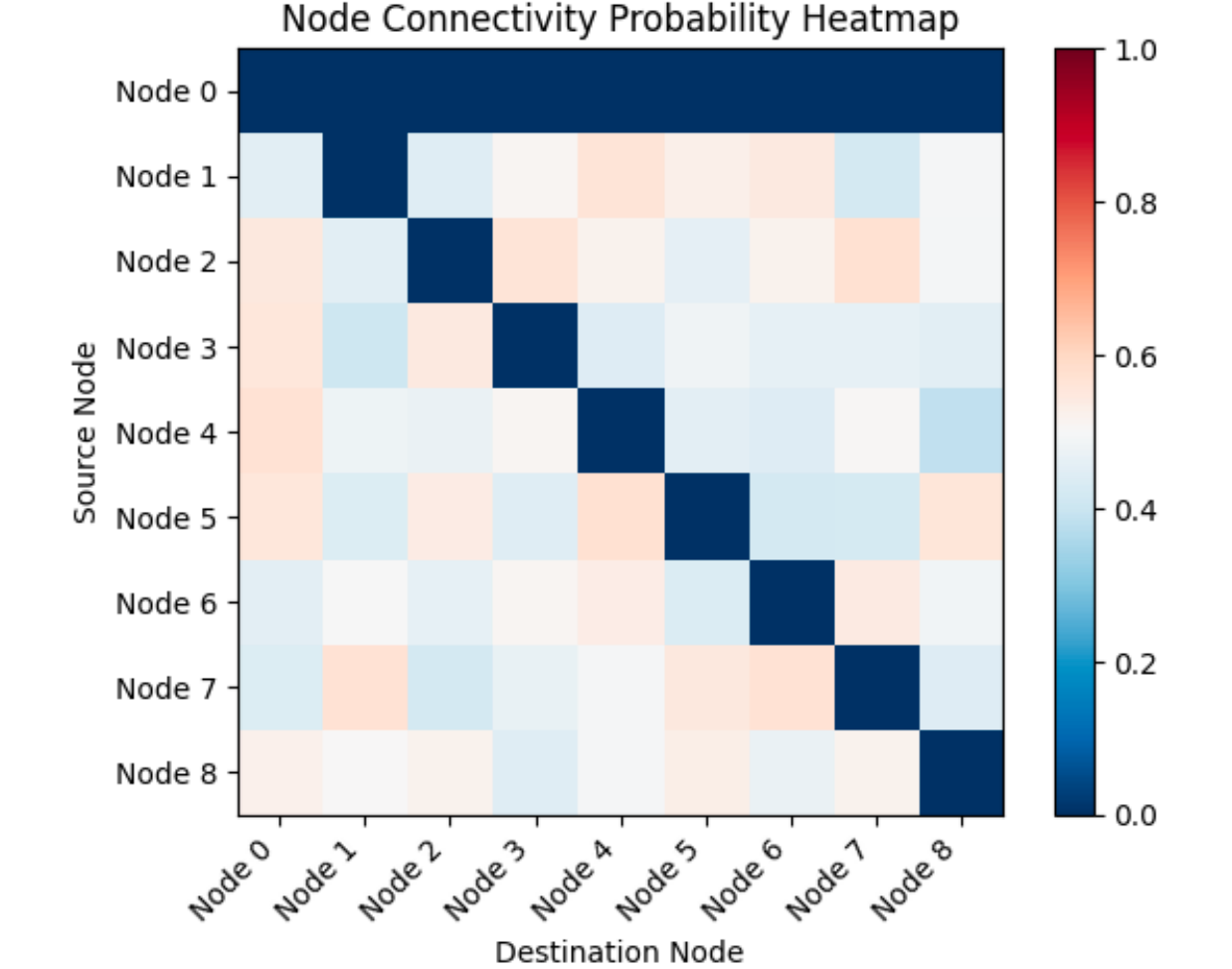} 
    \caption{Probabilities for sampling an edge in the graph by Dynamic Graph with the reduced model. Node 0 is the final decision node, nodes 1 to 4 are truthful agents using Gemma-7B-it, and nodes 5 to 8 are truthful agents using BlueLM-7B-chat. Notably, self-loops as well as connections from the final decision node to any other node are not allowed and thereby 0.}
    \label{fig:ablation_heatmap}
\end{figure}

\section{Mini Crosswords Puzzle Dataset}
\label{app: Mini Crosswords Puzzle}
We provide samples (Table \ref{tab:crosswords_sample}) of the Mini Crosswords Puzzle dataset used for the experiments in \ref{Crosswords Puzzle Experiment}.
\begin{table}[h!]
\centering
\caption{Sample Data from the Mini Crosswords Dataset}
\label{tab:crosswords_sample}

\begin{tabular}{|c|c|c|c|c|}
\hline
P & R & I & N & T \\ \hline
S & I & M & A & R \\ \hline
S & C & I & S & E  \\ \hline
S & E & N & S & E \\ \hline
T & R & E & A & D \\ \hline
\end{tabular}

\vspace{1cm} 

\begin{tabular}{|p{6cm}|c|}
\hline
\textbf{Clue} & \textbf{ID} \\ \hline
To stamp; to brand; to impress; to put into type & H1 \\ \hline
A scarf; a cymar; a loose dress & H2 \\ \hline
To cut & H3 \\ \hline
To perceive; wisdom; reason; feeling & H4 \\ \hline
The ridges on a tire; to walk heavily & H5 \\ \hline
A signaling sound & V1 \\ \hline
A rice processor; an implement for ricing & V2 \\ 
potatoes & \\ \hline
A chemical compound & V3 \\ \hline
A dog whelk or its shell & V4 \\ \hline
Chased up a tree & V5 \\ \hline
\end{tabular}
\end{table}

Below is the raw input formatted for clarity, \( \hookrightarrow \) indicates a line break:
\medskip

\begin{Verbatim}[frame=single, framesep=2mm, label=Input Data, fontsize=\small, breaklines=true]
[
    [
        "To stamp; to brand; to impress; to put into type",
        "A scarf; a cymar; a loose dress",
        "To cut",
        "To perceive; wisdom; reason; feeling",
        "The ridges on a tire; to walk heavily",
        "A signaling sound",
        "A rice processor; an implement for ricing potatoes",
        "A chemical compound",
        "A dog whelk or its shell",
        "Chased up a tree"
    ],
    [
        "P", "R", "I", "N", "T",
        "S", "I", "M", "A", "R",
        "S", "C", "I", "S", "E",
        "S", "E", "N", "S", "E",
        "T", "R", "E", "A", "D"
    ]
]
\end{Verbatim}

\section{MMLU Dataset}
\label{app:MMLU Dataset}
The following question is a data sample from the MMLU data set, used in the experiments in \ref{Adversarial Agent Detection} and \ref{Specialized Agents Experiment}. Specifically sampled from the test set category \textbf{College Mathematics}.

\begin{Verbatim}[frame=single, framesep=2mm, label=Dataset Question, fontsize=\small,breaklines=True]
Question: Let V and W be 4-dimensional subspaces of a 7-dimensional vector space X.
Which of the following CANNOT be the dimension of the subspace V intersect W?
Options:
A) 0
B) 1
C) 2
D) 3
Correct Answer: A
\end{Verbatim}

Below is the raw input formatted for clarity, \( \hookrightarrow \) indicates a line break:
\begin{Verbatim}[frame=single, framesep=2mm, label=Dataset Question, fontsize=\small,breaklines=True]
Let V and W be 4-dimensional subspaces of a 7-dimensional vector space X. Which of the following CANNOT be the dimension of the subspace V intersect W?,0,1,2,3,A
\end{Verbatim}

\section{CMMLU Dataset}
\label{app:CMMLU Dataset}
The following question is a data sample from the CMMLU data set, used in the experiments in \ref{Specialized Agents Experiment}. Specifically sampled from the test set category \textbf{Chinese Food Culture}.
\begin{CJK}{UTF8}{gbsn}
\begin{Verbatim}[frame=single, framesep=2mm, label=Dataset Question, fontsize=\small,breaklines=True]
Question: 传统名菜“松鼠桂鱼”是典型的什么菜？
Options:
A) 川菜
B) 粤菜
C) 淮扬菜
D) 鲁菜
Correct Answer: C
\end{Verbatim}
\end{CJK}

Below is the raw input formatted for clarity, \( \hookrightarrow \) indicates a line break:

\begin{CJK}{UTF8}{gbsn}
\begin{Verbatim}[frame=single, framesep=2mm, label=Input Data, fontsize=\small,breaklines=True]
传统名菜“松鼠桂鱼”是典型的什么菜？,川菜,粤菜,淮扬菜,鲁菜,C
\end{Verbatim}
\end{CJK}

In this appendix we provide samples for the prompts used in our experiments.
\section{MMLU Prompt Set}
\label{app:MMLU Prompt Set}
\DefineVerbatimEnvironment{PromptExample}{Verbatim}{
    fontsize=\small,
    frame=single,
    framesep=2mm,
    rulecolor=\color{black!30},
    fillcolor=\color{gray!5},
    breaklines=true,
}
This appendix provides detailed examples of prompts used in the study. These prompts are part of the \texttt{MMLUPromptSet} for a 4-option question answering framework. Except the constraint prompt, all prompts are adapted from \cite{zhuge2024language}.

\subsection{Role of the Prompt}
The role prompt is defined as follows:
\begin{PromptExample}
    a knowledgeable expert in question answering
\end{PromptExample}

\subsection{Constraints of the Prompt}
The constraints are outlined to ensure the response format and correctness. We changed the required output format to JSON format. This made parsing easier and helped the models to adhere better to a given format:
\begin{PromptExample}
"I will ask you a question. 
I will also give you 4 answers enumerated as A, B, C, and D.
Only one answer out of the offered 4 is correct. 
You must choose the correct answer to the question.
Answer with only a single letter (A, B, C, or D).
Do not include any other information in your answer except the letter.
Your response should be in JSON format, with the key 'answer' and the value being one of the 4 letters: A, B, C, or D, corresponding to the correct answer.
Here is an example of the correct format:
{
    'answer': 'A'
}"
\end{PromptExample}

\subsection{Formatting of the Response}
The expected format of the response is:
\begin{quote}
    one of the letters: A, B, C or D
\end{quote}
In JSON format.

\subsection{Example Prompts}
Here are some specific prompts used for different scenarios within the framework:

\subsection{Adversarial Answer Prompt}
Designed to receive a deceptive answer to the given question. We did not use this prompt for our experiments because the \glspl{LLM} seemed to be unable to output a lie:
\begin{PromptExample}
Answer a lie to the following question: {question}.
\end{PromptExample}

\subsection{Reflective Prompt}
Encourages reflection on the provided question and answer, assessing correctness and accuracy:
\begin{PromptExample}
Reflect on the following question and answer:
Question: {question}
Answer: {answer}
What are your thoughts on the correctness and accuracy of the answer? Do you agree or disagree? Why? Please provide a brief explanation.
\end{PromptExample}

\section{Crosswords Prompt Set}
\label{app:Crosswords Prompt Set}
\DefineVerbatimEnvironment{PromptExample}{Verbatim}{
fontsize=\small,
frame=single,
framesep=2mm,
rulecolor=\color{black!30},
fillcolor=\color{gray!5},
breaklines=true,
}
This appendix provides detailed examples of prompts used in the mini crossword puzzle game framework. These prompts are part of the \texttt{CrosswordsPromptSet}, designed to guide \glspl{LLM} in solving a 5 x 5 crossword puzzle. The prompts are mainly adapted from the framework and experiments by \citet{zhuge2024language}.
\subsection{Propose Prompt}
The prompt used for to start the game is the following:
\begin{PromptExample}
Let's play a 5 x 5 mini crossword, where each word should have exactly 5 letters.

{board}

Given the current status, list all possible answers for unfilled or changed words, and your confidence levels (certain/high/medium/low), using the JSON format with the position of the word as key and a list of lists consisting of possible answer and your confidence about the solution, as shown in this example {{"<position>": [["<answer>" , "<confidence>"]]}}. Use "certain" cautiously and only when you are 100\% sure this is the correct word. You can list more than one possible answer for each word. Each word should have a length of exactly 5 characters. Consider the intersection of horizontal and vertical words.
\end{PromptExample}

\subsection{Prompt to Test Correctness}
The prompt used to verify the correctness of an answer by an \gls{LLM}:
\begin{PromptExample}
"Does {word} has meaning "{meaning}"? Responde only Yes or No."
\end{PromptExample}

\subsection{Formatting of the Response}
The expected format of the response is:
\begin{quote}
a JSON object containing the position, possible answers, and their confidence levels
\end{quote}

\subsection{Suggest Prompt}
A prompt to inform an \gls{LLM} about a previous game and ask it to plan the next game:
\begin{PromptExample}
    You are playing a 5 x 5 mini crossword, where each word should have exactly 5 letters.
Given the current status: The target words are classified as {List of words and their categorization into "correct", "incorrect", and "impossible"} You will retry the game. Write a plan for the next time. 
Respond at most five sentences, one sentence per line.
Do not include the phrase "next time" in your response.
\end{PromptExample}

\subsection{Evaluation Prompt}
A prompt to make the \gls{LLM} evaluate the current state of the board and find possible solutions based on letters that have already been filled in.:
\begin{PromptExample}
    Evaluate if there exists a five letter word of some meaning that fit some letter constraints (sure/maybe/impossible).

Incorrect; to injure: w _ o _ g
The letter constraint is: 5 letters, letter 1 is w, letter 3 is o, letter 5 is g.
Some possible words that mean "Incorrect; to injure":
wrong (w r o n g): 5 letters, letter 1 is w, letter 3 is o, letter 5 is g. fit!
sure

A person with an all-consuming enthusiasm, such as for computers or anime: _ _ _ _ u
The letter constraint is: 5 letters, letter 5 is u.
Some possible words that mean "A person with an all-consuming enthusiasm, such as for computers or anime":
geek (g e e k): 4 letters, not 5
otaku (o t a k u): 5 letters, letter 5 is u
sure

Dewy; roscid: r _ _ _ l
The letter constraint is: 5 letters, letter 1 is r, letter 5 is l.
Some possible words that mean "Dewy; roscid":
moist (m o i s t): 5 letters, letter 1 is m, not r
humid (h u m i d): 5 letters, letter 1 is h, not r
I cannot think of any words now. Only 2 letters are constrained, it is still likely
maybe

A woodland: _ l _ d e
The letter constraint is: 5 letters, letter 2 is l, letter 4 is d, letter 5 is e.
Some possible words that mean "A woodland":
forest (f o r e s t): 6 letters, not 5
woods (w o o d s): 5 letters, letter 2 is o, not l
grove (g r o v e): 5 letters, letter 2 is r, not l
I cannot think of any words now. 3 letters are constrained, and _ l _ d e seems a common pattern
maybe

An inn: _ d _ w f
The letter constraint is: 5 letters, letter 2 is d, letter 4 is w, letter 5 is f.
Some possible words that mean "An inn":
hotel (h o t e l): 5 letters, letter 2 is o, not d
lodge (l o d g e): 5 letters, letter 2 is o, not d
I cannot think of any words now. 3 letters are constrained, and it is extremely unlikely to have a word with pattern _ d _ w f to mean "An inn"
impossible

Chance; a parasitic worm; a fish: w r a k _
The letter constraint is: 5 letters, letter 1 is w, letter 2 is r, letter 3 is a, letter 4 is k.
Some possible words that mean "Chance; a parasitic worm; a fish":
fluke (f l u k e): 5 letters, letter 1 is f, not w
I cannot think of any words now. 4 letters are constrained, and it is extremely unlikely to have a word with pattern w r a k _ to mean "Chance; a parasitic worm; a fish"
impossible

{input}
\end{PromptExample}

\section{CMMLU and MMLU Prompt Set}
\label{app:CMMLU and MMLU Prompt Set}
Direct translations of the prompts into Chinese were considered; however, such translations did not influence the performance outcomes of the models. Therefore, for simplicity and to streamline the implementation process, identical prompts were employed during training on both the \gls{MMLU} and \gls{CMMLU} datasets.

\end{document}